\documentclass[conference]{IEEEtran}
\pdfoutput=1
\usepackage{cite}
\usepackage{amsmath,amssymb,amsfonts}
\usepackage{algorithmic}
\usepackage{graphicx}
\usepackage{textcomp}
\usepackage{xcolor}
\usepackage{svg}
\usepackage{booktabs}
\usepackage{adjustbox}
\usepackage{float}
\usepackage[utf8]{inputenc} 
\usepackage[T1]{fontenc} 
\usepackage[english]{babel}
\usepackage{enumerate}
\usepackage[margin=0.8in]{geometry}
\usepackage{wrapfig}
\usepackage{enumitem,multirow}
\usepackage[final]{pdfpages}
\usepackage[none]{hyphenat}
\usepackage{color}
\usepackage{amsthm}
\usepackage{xcolor}
\usepackage{url}
\usepackage{dblfloatfix} 

\usepackage[normalem]{ulem}

\usepackage{todonotes}

\newcommand{\etal}{{\it et al.}}

\begin{document}

\bstctlcite{IEEEexample:BSTcontrol}

\title{Attention-based 3D Object Reconstruction \\ from a Single Image}



\author{\IEEEauthorblockN{Andrey Salvi\IEEEauthorrefmark{1}, Nathan Gavenski\IEEEauthorrefmark{1}, Eduardo Pooch\IEEEauthorrefmark{1}, Felipe Tasoniero\IEEEauthorrefmark{1} and Rodrigo Barros\IEEEauthorrefmark{2}}
\IEEEauthorblockA{School of Technology, Pontif{\'i}cia Universidade Cat{\'o}lica do Rio Grande do Sul \\
Av. Ipiranga, 6681, 90619-900, Porto Alegre, RS, Brazil}
\IEEEauthorrefmark{1}\{andrey.salvi, nathan.gavenski, eduardo.pooch, felipe.tasoniero\}@edu.pucrs.br,
\IEEEauthorrefmark{2}rodrigo.barros@pucrs.br
}

\maketitle

\begin{abstract} 
Recently, learning-based approaches for 3D reconstruction from 2D images have gained popularity due to its modern applications, e.g., 3D printers, autonomous robots, self-driving cars, virtual reality, and augmented reality. The computer vision community has applied a great effort in developing functions to reconstruct the full 3D geometry of objects and scenes. However, to extract image features, they rely on convolutional neural networks, which are ineffective in capturing long-range dependencies. In this paper, we propose to substantially improve Occupancy Networks, a state-of-the-art method for 3D object reconstruction. For such we apply the concept of self-attention within the network's encoder in order to leverage complementary input features rather than those based on local regions, helping the encoder to extract global information. With our approach, we were capable of improving the original work in 5.05\% of mesh IoU, 0.83\% of Normal Consistency, and more than $10\times$ the Chamfer-L1 distance. We also perform a qualitative study that shows that our approach was able to generate much more consistent meshes, confirming its increased generalization power over the current state-of-the-art.
\end{abstract}


\section{Introduction}
\footnotetext{Submitted to International Joint Conference on Neural Networks (IJCNN) 2020}
There are a variety of applications that make use of three-dimensional object models, like 3D printing, computer-generated imagery for scenario modeling on movies, object modeling for video-games, simulation of buildings in the field of architecture and civil engineering, and building reconstruction of archaeological sites. 
The current process of 3D modeling requires expert knowledge on modeling tools and techniques or specialized sensors for 3D reconstruction, such as contact methods (e.g., coordinate measuring machines) or non‐contact methods (e.g.,  X‐rays and laser scanning).
Still, there is no single modeling technique that satisfies every requirement of high geometric accuracy, portability, full automation, photo‐realism, low cost, flexibility, and efficiency \cite{remondino2006image}.  

One possible portable and low-cost solution is to model 3D objects based on simple commands, such as taking a picture of the object with a regular camera. There have been attempts to image-based 3D reconstruction by using geometrical measures of a sequence of images \cite{gruen2013calibration, horn1989shape} or of a single image \cite{van1998line, el2001flexible, remondino2003human}. These methods use hard-coded features, like image shading and texture, or human inputted features via interactive interfaces. 

Considering the success of deep neural networks approaches for function modeling~\cite{lecun2015deep}, current research on computer vision mostly revolves around Convolutional Neural Networks (CNNs) \cite{lecun1998gradient}.
CNNs have been successfully used for automated feature extraction of 2D images, achieving state-of-the-art in many computer vision tasks like image classification, object detection, and image captioning.
Reconstruction of 3D models from 2D images using artificial intelligence is an active area of research. 

Current work mostly relies on deep neural networks for feature extraction or structure prediction \cite{Wu_2015_CVPR, brock2016generative, Wu_etal_2016, Rezende_et_al_2016, Girdhar_etal_2016, choy2016, wang2018pixel2mesh, smith2018multi, smith2019geometrics, mescheder2019occupancy}.
Those applications propose to accelerate the creation of 3D models, automating the process and reducing the need of 3D modeling experts for simple modeling tasks so they can focus on refining the models, changing the way as different industries handle 3D modeling such as architecture, digital games, movies, and healthcare \cite{Yin_etal_2009, Guidi_etal_2014, Scheenstra_etal_2005}.

Most state-of-the-art methods on single image 3D reconstruction \cite{choy2016,  wang2018pixel2mesh, smith2018multi, smith2019geometrics, mescheder2019occupancy} exploit CNNs as feature extractors to capture relevant information from a given 2D image and then generate a 3D representation from the object. Roughly speaking, those methods generate the 3D surfaces in one of these three kinds of volume representation: 
\begin{itemize}
    \item Voxel: a regular grid representation of 3D surfaces in which we can infer the coordinates upon the relative position of a voxel to the others.
    \item Point Cloud: a cloud of points in the 3D space which represents the 3D surface. 
    \item Mesh: a 3D representation of a surface from an object using vertices (points in the 3D space), edges (connections between two vertices), and faces (closest set of edges).
\end{itemize}

\begin{figure*}[t]
    \centering
    \includegraphics[width=\linewidth]{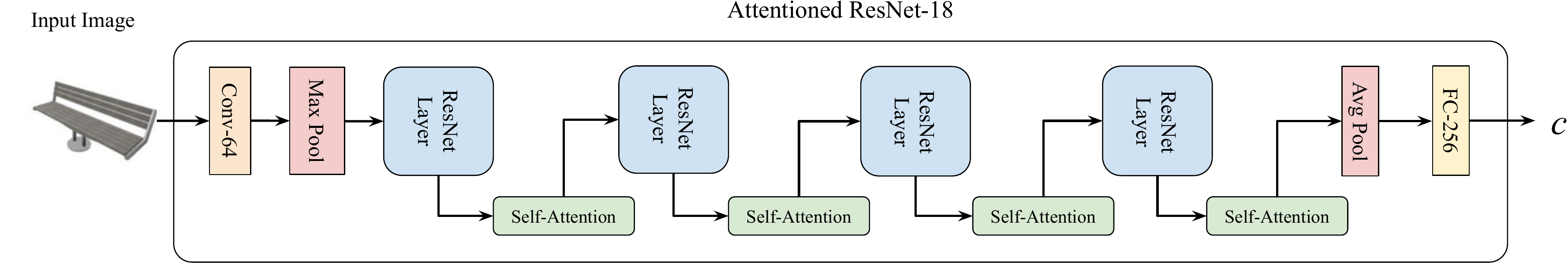}
    \caption{One of our proposed encoder architectures. This example is a ResNet-18 with four self-attention modules.}
    \label{fig:attn_encoder}
\end{figure*}

Meschender et al. \cite{mescheder2019occupancy} propose a novel method called Occupancy Networks (ONets) that represent 3D surfaces with a continuous decision boundary function, enabling extracting meshes at any resolution. ONet can reconstruct 3D objects from voxels, point clouds, and 2D images, achieving state-of-the-art results on 3D mesh reconstruction from 2D images on the ShapeNet dataset~\cite{chang2015shapenet}.

Despite those methods being capable of reconstructing a 3D volume from a 2D image, there are problems on the 3D reconstruction still poorly addressed, such as missing structural parts of the object (e.g., missing arms of an armchair), wrong textures (e.g., foiling a smooth texture), or nonexistent structures being added to the object (e.g., filling a leaked surface). Considering there is a single view angle from the object (based on the 2D image), the extracted features might not represent all the three-dimensional information (e.g., the unseen backside of the object). Since the convolution operation of a CNN works over local receptive fields, the network can only process long-range dependencies after many sequences of convolutional layers (i.e., a very deep architecture). Therefore, many CNNs fail to capture patterns on images across different image regions. We hypothesize that a mechanism such as self-attention~\cite{vaswani2017attention} could leverage the field of 3D model reconstruction from a single image, considering its previous success on machine translation~\cite{vaswani2017attention}, image generation~\cite{ZhangEtAl2019}, and other tasks in which one must capture global dependencies to succeed.

In this work, our main contribution is to substantially improve the Occupancy Networks~\cite{mescheder2019occupancy}, which outperformed previous approaches and is the current state-of-the-art on supervised single image 3D reconstruction. Our approach enhances ONet's encoder in order to extract more informative features from 2D images, and hence better model the latent feature space, and it does so by exploiting strategies that have been successful in other computer vision (and even natural language processing) tasks, such as self-attention and adaptive instance normalization.

\section{Related Work}
Reconstructing 3D objects from 2D images is an active research area in computer vision, and the interest in synthesizing 3D shapes with deep neural networks is increasing. Recent work in neural image synthesis has aimed at improving the fidelity of the resulting generated images with 3D-aware networks. 

Choy et al. \cite{choy2016} propose a recurrent neural network called 3D Recurrent Reconstruction Neural Network (3D-R2N2), which takes in one or more images of an object instance from different viewpoints to learn a reconstruction of the object in a 3D occupancy grid based on synthetic data in a supervised manner. For single-view image reconstruction, 3D-R2N2 achieved state-of-the-art on the ShapeNet dataset \cite{chang2015shapenet} at the time.

Wang et al. \cite{wang2018pixel2mesh} propose a supervised graph-based convolution algorithm that can extract a 3D triangular mesh from a single image. Their approach deforms an ellipsoid mesh with fixed size to the target geometry, allowing to refine the shape gradually, outperforming 3D-R2N2.

There are also unsupervised approaches to single image 3D reconstruction. Rezende~et~al.~\cite{rezende} propose a neural projection layer and a black-box renderer for supervising the learning process, which is built by first applying a transformation to the reconstructed volume, followed by a combination of 3D and 2D convolutional layers mapping the 3D volume into a 2D image. Yan~et~al.~\cite{yan2016} explore the task of 3D object reconstruction and proposes an encoder-decoder network that uses projection transformation as regularization, obtaining satisfactory performance in object reconstruction. Henderson~and~Ferrari \cite{henderson2018} present a unified framework for both reconstruction and generation of 3D shapes with only 2D supervision.

\section{Proposed Approach}

\subsection{Occupancy Networks}

The Occupancy Network \cite{mescheder2019occupancy} is composed of three modules: an initial encoder as a feature extractor, which can vary according to the input type (e.g., for 2D images the encoder is a ResNet-18 \cite{he2016deep}); a system with five Conditional Batch Normalization blocks as decoder of the generated features; and finally the occupancy function $o: \mathbb{R}^3 \rightarrow $\{0, 1\}, which classifies each point from the space whether or not it belongs to the surface.

The ResNet-18 encoder architecture contains four ResNet layers. Each ResNet layer consists of two ResNet blocks, and each ResNet block contains a convolutional layer followed by batch normalization, a ReLU activation function, and finally another convolutional layer followed by batch normalization. This process generates the features $c$ from the input image. We can see an example of ONet encoder with four blocks of self-attention in Figure~\ref{fig:attn_encoder}.

\begin{figure*}[t]
    \centering
    \includegraphics[width=0.9\linewidth]{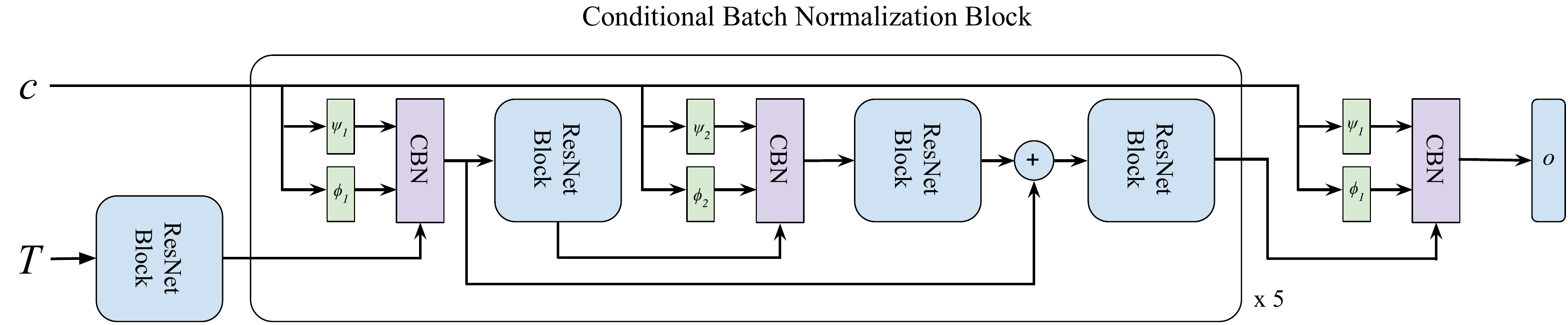}
    \caption{\label{fig:decoder} Decoder architecture. The decoder has five Conditional Batch Normalization Blocks.}
\end{figure*}

The decoder takes as input the features $c$ extracted from the encoder and a batch of learned 3D coordinates $T$. It is a system of five Conditional Batch Normalization (CBN) blocks, where each CBN block computes the batch of 3D coordinates via three ResNet blocks. Between each ResNet block, a CBN normalizes the tensor computed by the 3D coordinates over the features $c$ extracted from the 2D input. We can compute the CBN by passing the features $c$ through two parallel fully-connected layers $\phi(c)$ and $\psi(c)$ and then normalizing it as in Equation \ref{eq:CBN}:
\begin{equation}
	\label{eq:CBN}
	CBN(c, f_{in}) =\psi(c) \frac{f_{i n}-\mu}{\sqrt{\sigma^{2}+\epsilon}}+\phi(c)
\end{equation}
where $f_{in}$ is the tensor outputted by the previous ResNet block, $\mu$ and $\sigma$ is the mean and standard deviation over $f_{in}$. Figure~\ref{fig:decoder} shows a visual representation of the CBN blocks and the entire decoder. The CBN box on the diagram represents the operation defined in Equation~\ref{eq:CBN}.

Finally, ONet predicts the complete occupancy function $\textit{o}$ of the 3D object by approximating it with a neural network~\textit{f$_{\theta}$(p, k)} that given an observation $\textit{k} \in \textbf{$\chi$}$ as input assigns to every point $\textit{p} \in \mathbb{R}^3$ a probability to which it belongs to the object, as in Equation \ref{eq:onet}:

\begin{equation}
    \label{eq:onet}
    f_{\theta}: \mathbb{R}^{3} \times \mathcal{X} \rightarrow[0,1]
\end{equation}

ONet predicts all grid points as active (belongs to the object) if the output from ONet is greater than a threshold $\tau$.
Then, ONet divides the active voxels into eight subvoxels and re-evaluate them by the occupancy function $\textit{o}$, repeating this process iteratively until it reaches the desired resolution. The Marching Cubes algorithm \cite{Lorensen} is applied to the final resolution to extract an approximate isosurface. 

\subsection{Self-Attention} \label{sec:attention}
Our approach is to apply self-attention~\cite{vaswani2017attention} in ONet's encoder, as shown in the Figure \ref{fig:attn_encoder}, to focus on regions of interest on the image and generate meshes more related to the input object. This approach makes the network leverage complementary features rather than local regions, e.g., both arms of a chair. The self-attention module, when used in earlier layers, is also able to focus more on finer details (e.g., the fine details of a sofa), and when used in later layers, on structural features (e.g., not miss the rifle scope)~\cite{gatys2016image, karras2019style}.

With this understanding, we believe that applying the self-attention module makes our method more robust to reconstruct meshes. We expect to be able to correct different textures from the same object (not fill hollow spaces) and create more consistent objects without missing pieces (a sofa with no legs).

In our work, we use an attention module based on the Self-Attention Generative Adversarial Network (SAGAN)~\cite{ZhangEtAl2019}. SAGAN uses a self-attention module over internal network states, outperforming prior work in image synthesis. The self-attention module calculates response at a position as a weighted sum of the features at all positions, capturing global dependencies with a small computational cost~\cite{zhang2018self}.

Given an input feature map $z$ from a previous layer, first we compute the key $f(z)$, the query $g(z)$ and the value $h(z)$ with convolutional filters of size 1x1 and the equations $f(z) = W_fz$, $g(z) = W_gz$, and $h(z) = W_hz$. With the key $f$ and the query $g$, we can compute the attention map in two steps. The first step is shown Equation \ref{eq_sij},

\begin{equation}
    \label{eq_sij}
    s_{ij} = f(z_i)^T g(z_j)
\end{equation}
then, we compute the softmax function $\beta_{j,i}$ over the $s_{ij}$, which indicates the network attention to the $i$-th location when synthesizing the $j$-th region. With the attention map $\beta$ and the values $h(z)$, now we can compute the self-attention feature maps $a = (a_1, a_2, ..., a_N) \in \mathbb{R}^{C \times N}$ as shown in Equation \ref{eq_o},
\begin{equation}
    \label{eq_o}
    a_j = \upsilon \left( \sum_{i=1}^{N} \beta _{j, i} h(z_i) \right), \\ \upsilon(z_i) = W_\upsilon z_i
\end{equation}
where $N$ is the number of feature locations and $C$ is the number of channels. In this formulation, $W_f$, $W_g$ and $W_h \in \mathbb{R}^{\tilde{C} \times C}$ and $W_\upsilon \in \mathbb{R}^{C \times \tilde{C}}$, where $\tilde{C}$ is $C/k$ to reduce the number of features. 

After computing the self-attention feature map $a$, we perform a normalization operation to compute the final output as shown in Equation \ref{eq_y},
\begin{equation}
    \label{eq_y}
    y_i = \gamma a_i + z_i
\end{equation}
where $\gamma$ is a learnable parameter initialized as 0.

\subsection{Ensemble Approach}
In our previous experiments, we observed that some models trained with self-attention just in one category outperform the model trained in all the categories, showing that the high diversity of objects in all the dataset does not improve the model in terms of generalization. Based on these preliminary results, we propose an ensemble of ONets, where each category has one specialized ONet. 

\begin{figure*}[t]
    \centering
    \includegraphics[width=\textwidth]{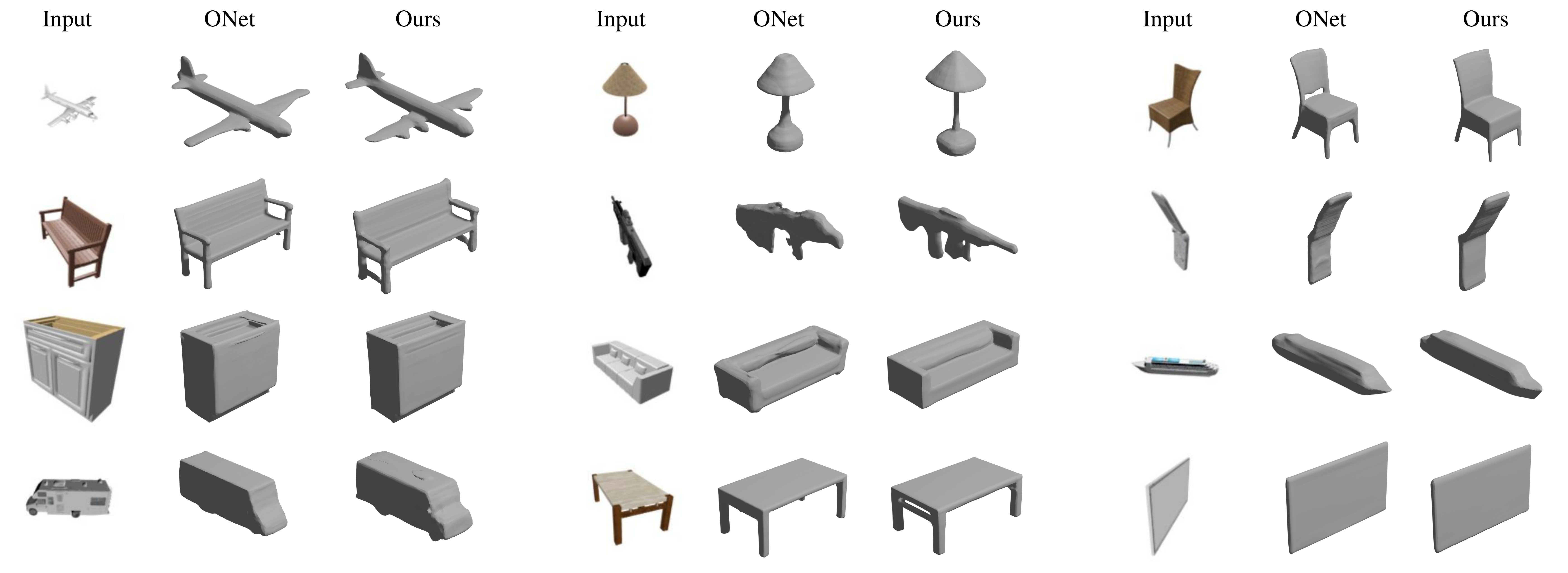}
    \caption{\label{fig_shapenet} Generated meshes based on a 2D image input. Comparison between ONet \cite{mescheder2019occupancy} and our approach when receiving the same input. Showing results for 12 different ShapeNet classes}
\end{figure*}

To create this ensemble of ONets, we evaluate three ONet versions for each category, and we use the best model in each category. The first version is an original ONet. The second version is an ONet attentioned in the initial layers, which were shown to represent details of high granularity \cite{ZhangEtAl2019}. The attention modules are added after the first and second ResNet layers. This means that the neurons have a small receptive field, looking over a smaller area on the input, and the attention will impact on the general structure of the output. The third version is an ONet attentioned in the last layers, representing details of small granularity. The attention modules are added after the third and fourth ResNet layers, meaning that the neurons have a large receptive field, looking over a larger area of the input, and the attention will impact on the finer details of the output.

In our final approach, ONets have self-attention modules after the first and second ResNet layers, except for the bench, display, and lamp categories, which have self-attention over the third and fourth ResNet layers.

\section{Experiments}

\subsection{Data}
To train and evaluate our model both quantitatively and qualitatively, we use the ShapeNet dataset \cite{chang2015shapenet} subset of Choy \etal~\cite{choy2016}. This subset consists of images of thirteen classes of objects, with 24 images of different viewpoints of each object in a 137x137 resolution, and for each object in the dataset, it has the expected surface in meshes, point clouds, and voxels. We use the official train/test split.

To test for generalization, we also use a subset of Stanford Online Products Dataset \cite{Song_2016_CVPR}. 
This dataset presents real images,obtained from online products available on eBay. Unlike ShapeNet, which presents synthetic data, Online Products contains pictures of real images.
The dataset contains 120,053 images of 22,634 products (classes) with different resolutions. It was originally used for image retrieval. 
In this work, we use this dataset to evaluate the models just in a qualitative manner, since this dataset was not created for the task of 3D reconstruction and does not contain ground-truth meshes or voxels to evaluate the quality of generated volume quantitatively. 
The subset used to perform single image 3D reconstruction are the cabinet, chair, lamp, sofa, and table product categories, which are the common objects between ShapeNet and Stanford Online Products.

\subsection{Metrics}
We evaluate our proposed method with the same metrics as Mescheder \textit{et al.} \cite{mescheder2019occupancy}: the Intersection Over Union (IoU) of the generated mesh with the ground-truth, the Chamfer-L1 distance, and the Normal Consistency.

\subsubsection{IoU}
The Intersection over Union (IoU) of the generated mesh with the ground-truth is computed as the quotient of the volume of the two meshes union and the volume of their intersection. 
The intersection is computed by randomly sampling 100,000 points from the bounding volume and determining if the predicted points lie inside or outside the ground truth.

\subsubsection{Chamfer-L1}
Chamfer distance is a metric to measure the distance between two edges of images/volumes, where a set of points represents the edges \cite{gesture-based}. Given two images $X$ and $Y$, the Equation \ref{chamfer_distance} computes the Chamfer distance 

\begin{equation}
    \label{chamfer_distance}
    d_{Chamfer}\left(X, Y\right)=\sum_{x \in X} \min _{y \in Y}\|x-y\|_{2}^{2}+\sum_{y \in Y} \min _{x \in X}\|x-y\|_{2}^{2}
\end{equation}
where $\|x-y\|_{2}^{2}$ is the Euclidean distance between two points, $x$ and $y$, belonging to the 2D objects $X$ and $Y$ respectively. 
The Chamfer distance penalizes for points belonging to the edge of $X$ that are so far from any point in the edge of $Y$. Also, it does not obey the triangle inequality rule.

\begin{table*}[!t]
\scriptsize
\centering
\caption{Quantitative results of Single Image 3D Reconstruction. Results from 3D-R2N2, Pix2Mesh and Onet from \cite{mescheder2019occupancy}.}
\label{quantitative_results}
\begin{tabular*}{\textwidth}{@{\extracolsep{\fill}}c|rrrr|rrrr|rrrr}
\toprule
 & \multicolumn{4}{c}{$IoU$} & \multicolumn{4}{c}{$Chamfer L-1$} & \multicolumn{4}{c}{$Normal\: Consistency$} \\
 Category & 3D-R2N2 & Pix2Mesh & Onet & Ours & 3D-R2N2 & Pix2Mesh & Onet & Ours & 3D-R2N2 & Pix2Mesh & Onet & Ours \\
\toprule
airplane    & $0.426$ & $0.420$ & $0.571$ & $\mathbf{0.645}$ & $0.227$ & $0.187$ & $0.147$ & $\mathbf{0.011}$ & $0.629$ & $0.759$ & $0.840$           & $\mathbf{0.868}$ \\
bench       & $0.373$ & $0.323$ & $0.485$ & $\mathbf{0.493}$ & $0.194$ & $0.201$ & $0.155$ & $\mathbf{0.016}$ & $0.678$ & $0.732$ & $\mathbf{0.813}$  & $\mathbf{0.813}$ \\
cabinet     & $0.667$ & $0.664$ & $0.733$ & $\mathbf{0.737}$ & $0.217$ & $0.196$ & $0.167$ & $\mathbf{0.016}$ & $0.782$ & $0.834$ & $\mathbf{0.879}$  & $0.876$ \\
car         & $0.661$ & $0.552$ & $0.737$ & $\mathbf{0.761}$ & $0.213$ & $0.180$ & $0.159$ & $\mathbf{0.014}$ & $0.714$ & $0.756$ & $0.852$           & $\mathbf{0.855}$ \\
chair       & $0.439$ & $0.396$ & $0.501$ & $\mathbf{0.534}$ & $0.270$ & $0.265$ & $0.228$ & $\mathbf{0.021}$ & $0.663$ & $0.746$ & $0.823$           & $\mathbf{0.829}$ \\
display     & $0.440$ & $0.490$ & $0.471$ & $\mathbf{0.520}$ & $0.314$ & $0.239$ & $0.278$ & $\mathbf{0.026}$ & $0.720$ & $0.830$ & $0.854$           & $\mathbf{0.863}$ \\
lamp        & $0.281$ & $0.323$ & $0.371$ & $\mathbf{0.379}$ & $0.778$ & $0.308$ & $0.479$ & $\mathbf{0.045}$ & $0.560$ & $0.666$ & $\mathbf{0.731}$  & $0.722$ \\
loudspeaker & $0.611$ & $0.599$ & $0.647$ & $\mathbf{0.660}$ & $0.318$ & $0.285$ & $0.300$ & $\mathbf{0.028}$ & $0.711$ & $0.782$ & $0.832$           & $\mathbf{0.839}$ \\
rifle       & $0.375$ & $0.402$ & $0.474$ & $\mathbf{0.527}$ & $0.183$ & $0.164$ & $0.141$ & $\mathbf{0.012}$ & $0.670$ & $0.718$ & $0.766$           & $\mathbf{0.804}$ \\
sofa        & $0.626$ & $0.613$ & $0.680$ & $\mathbf{0.689}$ & $0.229$ & $0.212$ & $0.194$ & $\mathbf{0.019}$ & $0.731$ & $0.820$ & $0.863$           & $\mathbf{0.866}$ \\
table       & $0.420$ & $0.395$ & $0.506$ & $\mathbf{0.535}$ & $0.239$ & $0.218$ & $0.189$ & $\mathbf{0.019}$ & $0.732$ & $0.784$ & $0.858$           & $\mathbf{0.861}$ \\
telephone   & $0.611$ & $0.661$ & $0.720$ & $\mathbf{0.754}$ & $0.195$ & $0.149$ & $0.140$ & $\mathbf{0.012}$ & $0.817$ & $0.907$ & $0.935$           & $\mathbf{0.937}$ \\
vessel      & $0.482$ & $0.397$ & $0.530$ & $\mathbf{0.568}$ & $0.238$ & $0.212$ & $0.218$ & $\mathbf{0.018}$ & $0.629$ & $0.699$ & $0.794$           & $\mathbf{0.801}$ \\
\midrule
mean        & $0.493$ & $0.480$ & $0.571$ & $\mathbf{0.600}$ & $0.278$ & $0.216$ & $0.215$ & $\mathbf{0.019}$ & $0.695$ & $0.772$ & $0.834$           & $\mathbf{0.841}$ \\
\bottomrule
\end{tabular*}
\end{table*}

The Chamfer distance has a high computational cost for meshes due to the high number of points, so it is interesting to compute an approximation \cite{gesture-based}. For this purpose, we use the Chamfer-L1, an approximation using $L_1$ norm as in \cite{mescheder2019occupancy}, by the Equation \ref{chamfer_L1}:

\begin{equation}
    \label{chamfer_L1}
    \begin{split}
     \text{Chamfer}L_1 (\mathcal{M} _ { \mathrm { pred } } , \mathcal { M } _ { \mathrm { GT } } )  \equiv \\
     \frac{1}{ 2 \left| \partial \mathcal{M}_{\mathrm{pred}} \right| } \int_{\partial \mathcal{M}_{\mathrm{pred}} \in \partial \mathcal{M}_{\mathrm{Or}}}\|p-q\| \mathrm{d} p + \\
     \frac{1}{ 2 \left| \partial \mathcal{M}_{\mathrm{GT}} \right| } \int_{\partial \mathcal{M}_{\mathrm{GT}} p \in \partial \mathcal{M}_{\mathrm{red}}}\|p-q\| \mathrm{d} q
    \end{split}
\end{equation}
where $\mathcal{M}_{\mathrm{pred}}$ and $\mathcal{M}_{\mathrm{GT}}$ are the meshes from a prediction and the ground-truth, respectively, and $\partial \mathcal{M}_{\mathrm{pred}}$ $\partial \mathcal{M}_{\mathrm{GT}}$ represents the surfaces of the two meshes. 
As in the work of Mescheder \etal~\cite{mescheder2019occupancy}, we sample 100,000 random points to represent the surface. 
Chamfer-L1 is a dissimilarity metric, so lower values mean more favorable results.

\subsubsection{Normal Consistency}
We compute the Normal Consistency over two meshes by the Equation \ref{eq:normal},
\begin{equation}
    \label{eq:normal}
    \begin{split}
    \text{NormalConsistency} \left(\mathcal{M}_{\mathrm{pred}}, \mathcal{M}_{\mathrm{GT}}\right) \equiv \\
    \frac{1}{2\left|\partial \mathcal{M}_{\mathrm{pred}}\right|} \int_{\partial \mathcal{M}_{\mathrm{pred}}}\left|\left\langle n(p), n\left(\mathrm{proj}_{2}(p)\right)\right\rangle\right| \mathrm{d} p + \\
    \frac{1}{2\left|\partial \mathcal{M}_{\mathrm{GT}}\right|} \int_{\partial \mathcal{M}_{\mathrm{GF}}}\left|\left\langle n\left(\mathrm{proj}_{1}(q)\right), n(q)\right\rangle\right| \mathrm{d} q
    \end{split}
\end{equation}
where $\langle\cdot, \cdot\rangle$ is the inner product over two vectors, $n(p)$ and $n(q)$ are the normal vectors on the meshes $\partial \mathcal{M}_{\mathrm{pred}}$ and $\partial \mathcal{M}_{\mathrm{GT}}$ and $proj_2(p)$ and $proj_1(q)$ denote the projections of $p$ and $q$ onto the meshes of ground-truth and prediction, respectively.

This metric is a way to measure how two volumes are consistent between them. For instance, two meshes may have a high IoU belonging to different classes. The consistency will measure the sharp differences between the objects and penalize the score in a way that does not occur in the IoU.

\subsection{Implementation Details}

In our work, we train our models using Adam Optimizer \cite{kingma2014adam}, with a learning rate of 0.001, weight decay of 1e-5, and the default betas $\beta _1$ of 0.9 and $\beta _2$ of 0.999. We train for 200K steps, evaluating the validation subset at every 2,000 steps and saving the models that achieve the best validation loss. In other ONet hyperparameters, we use the default from Mescheder et al.'s implementation \cite{mescheder2019occupancy}.

\section{Results} \label{sec:results}

\subsection{Quantitative Results}


As shown in Table \ref{quantitative_results}, our proposed approach performs slightly better in all object categories using the IoU metric. Comparing the Normal Consistency, our model's performance is better in most categories, except in cabinet and lamp. We achieve the most significant results comparing the methods with the Chamfer-L1 distance measure. On average, our approach improves ONet's IoU by 5.05\%, the Chamfer-L1 decreases more than 10 times, and the Normal Consistency improves in 0.83\%. Figure \ref{fig_shapenet} shows some 3D reconstruction results on different ShapeNet categories.  The most significant difference that we observed was in the rifle category, in which our method generates images with fewer deformations and more accurate fine details. The airplane category also show more detailed structural components.

In Table \ref{quantitative_results_2}, we show the results of training three ONets per category. The first ONet does not have self-attention, the second has self-attention after inital ResNet layers, and the third ONet has self-attention after final ResNet layers. All the models achieve a mean IoU, Chamfer-L1, and Normal Consistency higher than the original ONet. In general, the models with attention on final layers achieve better results, resulting in higher IoU and Normal Consistency and similar Chamfer-L1 to the model without attention.

\subsection{Qualitative Results}

In this experiment, we perform single image 3D reconstruction using the models trained on the ShapeNet train set and evaluate these models both on ShapeNet test set and a subset of Stanford Online Products to validate the models' generalization power.

\begin{table*}[!t]
\scriptsize
\centering
\caption{Quantitative results of the Onets trained per category. w.o. attn, attn 1-2, attn 3-4.}
\label{quantitative_results_2}
\begin{tabular*}{\textwidth}{@{\extracolsep{\fill}}c|ccc|ccc|ccc}
\toprule
& \multicolumn{3}{c}{$IoU$} & \multicolumn{3}{c}{$Chamfer-L1$} & \multicolumn{3}{c}{$Normal\: Concistency$} \\
category & w.o. attn & attn 1-2 & attn 3-4 & w.o. attn & attn 1-2 & attn 3-4 & w.o. attn & attn 1-2 & attn 3-4  \\
\toprule
airplane    & $\mathbf{0.649}$  & $0.645$           & $0.637$           & $\mathbf{0.010}$  & $0.011$           & $0.011$           & $\mathbf{0.868}$  & $\mathbf{0.868}$  & $0.865$ \\
bench       & $0.434$           & $0.461$           & $\mathbf{0.493}$  & $0.017$           & $\mathbf{0.016}$  & $\mathbf{0.016}$  & $0.806$           & $\mathbf{0.813}$  & $\mathbf{0.813}$ \\
cabinet     & $0.736$           & $\mathbf{0.737}$  & $0.732$           & $0.017$           & $0.017$           & $\mathbf{0.016}$  & $0.873$           & $\mathbf{0.876}$  & $0.873$ \\
car         & $\mathbf{0.761}$  & $\mathbf{0.761}$  & $0.756$           & $\mathbf{0.013}$  & $0.014$           & $0.014$           & $0.854$           & $\mathbf{0.856}$  & $0.855$ \\
chair       & $0.531$           & $\mathbf{0.534}$  & $0.530$           & $\mathbf{0.021}$  & $\mathbf{0.021}$  & $0.022$           & $0.828$           & $\mathbf{0.829}$  & $0.828$ \\
display     & $0.515$           & $0.516$           & $\mathbf{0.520}$  & $\mathbf{0.026}$  & $0.027$           & $\mathbf{0.026}$  & $0.857$           & $\mathbf{0.864}$  & $0.863$ \\
lamp        & $\mathbf{0.384}$  & $0.377$           & $0.379$           & $0.040$           & $\mathbf{0.038}$  & $0.045$           & $0.734$           & $\mathbf{0.739}$  & $0.722$ \\
loudspeaker & $0.657$           & $\mathbf{0.660}$  & $0.647$           & $\mathbf{0.027}$  & $0.028$           & $0.030$           & $0.835$           & $\mathbf{0.839}$  & $0.836$ \\
rifle       & $0.520$           & $\mathbf{0.527}$  & $0.518$           & $\mathbf{0.012}$  & $\mathbf{0.012}$  & $\mathbf{0.012}$  & $0.802$           & $\mathbf{0.804}$  & $0.800$ \\
sofa        & $0.684$           & $\mathbf{0.689}$  & $0.687$           & $\mathbf{0.019}$  & $\mathbf{0.019}$  & $\mathbf{0.019}$  & $0.862$           & $\mathbf{0.866}$  & $0.865$ \\
table       & $0.530$           & $\mathbf{0.535}$  & $0.532$           & $\mathbf{0.018}$  & $0.019$           & $0.019$           & $0.860$           & $\mathbf{0.861}$  & $0.860$ \\
telephone   & $0.743$           & $\mathbf{0.754}$  & $0.745$           & $0.013$           & $\mathbf{0.012}$  & $0.013$           & $0.937$           & $\mathbf{0.939}$  & $0.937$ \\
vessel      & $0.565$           & $\mathbf{0.568}$  & $0.564$           & $0.019$           & $\mathbf{0.018}$  & $0.019$           & $0.800$           & $0.801$           & $\mathbf{0.802}$ \\
\midrule
mean        & $0.593$           & $\mathbf{0.597}$ & $0.595$            & $\mathbf{0.019}$  & $\mathbf{0.019}$  & $0.020$           & $0.840$           & $\mathbf{0.843}$  & $0.840$ \\
\bottomrule
\end{tabular*}
\end{table*}

As opposed to Mescheder \textit{et al.} \cite{mescheder2019occupancy}, we do not segment object regions in our qualitative results, and we do not retrain the model in other view angles. As Online Products contains images with background, we also it to evaluate the results in a more realistic scenario.

To evaluate qualitatively, we created an online survey \footnote{The form is available at https://forms.gle/mGmwJ3vuTFLpFPcT8} with the question: "Which image makes a more accurate 3D representation of the original image?". The form shows the input image and the two outputs generated by ONet and by our proposed method to the user, with two radio buttons to user answer the question. There are a total of 25 images, five per category. We block the users from making multiple answers and we shuffle the alternatives to prevent bias.

At the time of this writing, 130 people answered our form. Our method achieves a mean of 117.68 votes as the most consistent  output with the input against a mean of 12.32 of the ONet approach (the standard deviation was 15.49).  In general, our approach achieved more than 90\% of the votes. 

\begin{figure*}[!t]
    \centering
    \includegraphics[width=0.7\textwidth]{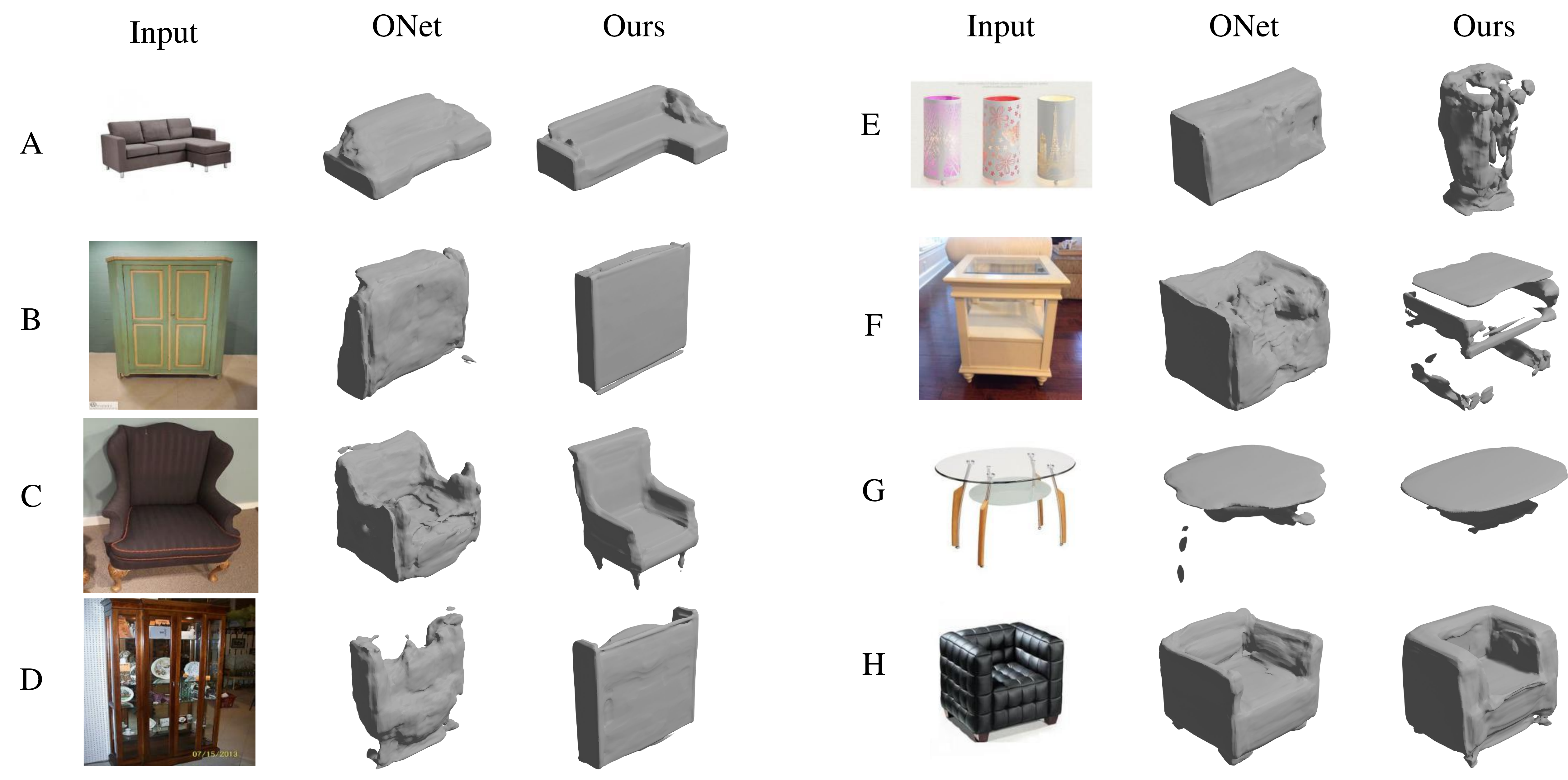}
    \caption{\label{fig:online_examples}Examples of 3D reconstruction on Stanford's Online Products  \cite{Song_2016_CVPR}}
\end{figure*}

We can see some exceptional cases in Figure \ref{fig:online_examples}. In the Input \textbf{A}, our mesh achieved 100\% of the votes. The self-attention module allows our model to generate the sofa arms and the L-shape, unlike the original ONet. In the Input \textbf{E}, our approach achieved 79 votes against 43 from ONet. ONet predicts a more filled mesh, however, with a squared shape. Our approach can not fill the mesh but generates in a more cylindrical shape, like the lamps in the input. It is important to note that this picture contains three objects, and all the images from ShapeNet contain just one object per image. In Input \textbf{F}, our approach achieved 78 votes against 54 from ONet. ONet generates a solid block with some deformations, more similar to a chair. However, the input is an end table, which haves a space between the upper and lower structures. Out method partially generate these structures, but it missed to connect them. In the Input \textbf{G}, our approach achieves 84 votes against 48. Both models generate the table top, and ONet even generates pieces from a table leg. Our approach missed the table leg, but generated the table in a rounder and smoother shape as the input. In the Input \textbf{H}, our approach achieved 103 votes against 29 from ONet. Both models generate meshes consistent with the input. ONet generates the chair feet more similar to the input, and the mesh is smoother than ours. In our approach, the self-attention allows generating thicker sofa arms, square shape, and the chair rests lower, as it is in the input.

In general, we can see in Figure \ref{fig:online_examples} that both meshes are similar to the inputs \textbf{A}, \textbf{C}, \textbf{G}, and \textbf{H}. Our approach generates meshes clearly more consistent with the inputs \textbf{B}, \textbf{C}, and \textbf{D}, cases where the images have a background, showing that our approach managed to generalize comparably well in real data.

\subsection{Ablations}

In this section, we briefly discuss some different approaches results and their performance (shown in Table \ref{quantitative_results3}). 

\subsubsection{Specialization} 
ONet method consists of training the model with all images at once.
We use the same approach to test whether using a self-attention module for all classes would help our model as it benefits the original method.
However, training our method with all classes wield inferior results.
When using the self-attention module in earlier layers, our model achieves $IoU$ of $0.575$, when using it in the later layers $0.587$, and after all ResNet blocks $0.568$.
We believe that those results are consequences of high variability between objects structures, which makes it harder to learn weights that properly weight all the different ShapeNet categories.

By creating a model for all classes, ONet enables the learning phase to be executed only once and a single model to be responsible for all meshes.
Nevertheless, by trying to fit all model's weights for all objects, the original method lacks representing fine details, as observed in Figure \ref{fig_shapenet}.
For that reason, we split all possible classes in different models, reducing the complexity of the learning curve by enabling the model to specialize in one single type of object.

We compare our results with both the ONet baseline and by retraining the original method for a single object as well.
When training ONet for each category (w.o. attn), the model achieves better results than the original, due to the specialization as we expect. 
The specialized model increases $IoU$ in $0.022$ and Normal Consistency in $0.006$, and decreases Chamfer-L1 in $0,196$. 
As for the model with the self-attention modules in the earlier layers (attn 1-2), we observed an increase of $0.029$ in $IoU$, $0.009$ in Normal Consistency, and the same result as the specialized model on Chamfer-L1. 
We believe that this improvement is due to the structural importance of the classes that the self-attention module provides.
Cars, chairs, loudspeakers, sofas, tables, telephones, and vessels do have meshes with much harder structural information to learn.
In cases where an object presents an unusual shape, e.g., the "L" shaped sofa in Figure \ref{fig:online_examples}, our model has a much easier time reproducing its mesh than the one without attention.
Meanwhile, displays and benches do have smaller details that the later attention modules tend to pay attention to.
The later attention modules (attn 3-4) help the model create more detailed meshes, e.g., the bezels of the displays.
Using the attention in the later layers increases $IoU$ in $0.020$ and Normal Consistency in $0.005$, and decreases Chamfer-L1 in $0,193$.

We understand that even though having more than one model might not be ideal since it depends on the model to know from which class the original image belongs. 
However, we believe that the trade-off given the qualitative results is worth it, since we can solve this problem on inference time with an image classification neural network, classifying input images and selecting the best model for each input.


\subsubsection{Normalization} 
We changed the Conditional Batch Normalization used by Mescheder \textit{et al.} \cite{mescheder2019occupancy} with the Adaptive Instance Normalization (AdaIN) from Zhang \textit{et al.} \cite{ZhangEtAl2019}. We were motivated to do that since Zhang \etal \cite{ZhangEtAl2019} uses AdaIN and self-attention to improve their experiments and their encoder ONet's encoder have a similar structure. This model achieves a mean of 0.579 in mesh IoU, and despite outperforming the original ONet in mesh IoU by 0.009, the model achieves a normal consistency of 0.622, underperforming the original ONet by 0.212. We observed that the model performed very well in cars and tables, but deforms objects from the class monitor and consumed pieces from lamp objects. Despite this behavior, these results indicate that AdaIN might be useful in this scenario and could be further investigated in future work.

\begin{table}[H]
\scriptsize
\centering
\caption{Influence of AdaIN and another self-attentions.}
\label{quantitative_results3}
    \begin{tabular*}{\columnwidth}{@{\extracolsep{\fill}}lc}
        \toprule
            Category & $IoU$ \\
        \toprule
            AdaIN                   & $0.579$ \\
            single attn 1-2         & $0.574$ \\
            single attn 3-4         & $0.587$ \\
            single attn 1-2-3-4     & $0.567$ \\
        \bottomrule
    \end{tabular*}
\end{table}

\subsubsection{Feature Extraction} 
We also tried changing the Resnet-18 encoder from ONet by the generator network of HoloGAN from the work of Thu \etal \cite{hologan} as a feature extractor from the input images. Our motivation for this experiment was the ability of HoloGAN to generate images of an object from angles never seen by the model before. Thu \etal \cite{hologan} train HoloGAN also on ShapeNet, bringing us the idea that the generator network from HoloGAN has a better power of imagination from the not seen sides of object that can help the ONet. Since Thu et al. train HoloGAN in a not supervised manner, they generate the images from a latent space feature, randomly sampled from a uniform distribution. To use the generator from HoloGAN to extract features from the objects, we also need to train a third model to learn the latent space features. This model was a VGG-19 that receives the input image, creates the features of the latent space. HoloGAN receives as input these latent space features and generates the features from the object that finally feeds on the ONet. In these experiments, the model does not learn correctly, generating meaningless meshes.

\section{Conclusions}

In this paper, we introduce a new approach employing the self-attention mechanism to improve ONet performance on single image 3D object reconstruction. Our experiments show that the self-attention mechanism has better results if trained separately for each object category. This approach allows the model to generate more consistent meshes in images of real objects and images with varying backgrounds, showing that the attention mechanism helped the model to ignore the irrelevant details of the image. Our approach improves previous approaches results, both quantitatively and qualitatively. Our method was able to generate more consistent meshes in real data, even though we trained it using synthetic data, showing that our approach can generalize to other domains. 

We believe that applying the self-attention mechanism also in the decoder will significantly increase our model performance. However, this experiment is computationally impractical, since it generates a tensor with 323 GB of memory in the self-attention execution. Investing a way to work around this problem is a challenge for future work.

\section*{Acknowledgment}
This paper was achieved in cooperation with HP Brasil Ind\'ustria e Com\'ercio de Equipamentos Eletr\^onicos LTDA. using incentives of Brazilian Informatics Law (Law n\textsuperscript{\underline{o}} 8.2.48 of 1991).

\bibliographystyle{IEEEtran}
\bibliography{references}


\end{document}